# Comparing K-Nearest Neighbors and Potential Energy Method in classification problem. A case study using KNN applet by E.M. Mirkes and real life benchmark data sets

Yanshan Shi
University of Leicester
ys98@leicester.ac.uk

**Abstract:** *K-nearest neighbors (KNN) method is used in many supervised learning classification problems. Potential Energy (PE) method is also developed for classification problems based on its physical metaphor. The energy potential used in the experiments are Yukawa potential and Gaussian Potential. In this paper, I use both applet and MATLAB program with real life benchmark data to analyze the performances of KNN and PE method in classification problems. The results show that in general, KNN and PE methods have similar performance. In particular, PE with Yukawa potential has worse performance than KNN when the density of the data is higher in the distribution of the database. When the Gaussian potential is applied, the results from PE and KNN have similar behavior. The indicators used are correlation coefficients and information gain.*

1.  **Introduction**

The target of supervised learning is to learn a mapping from the input to an output whose correct values are provided. However for unsupervised learning, no correct values are provided hence the only known object is the input data and the target is to find the regularities in the input.

Classification is considered as an object of supervised learning. It is the problem to identify the category that a new observation belongs given a training set of data. In this training set, all data are labeled with their corresponding categories. The algorithm used for classification purpose is named the classifier. In this paper, I compare the performance of two classifiers: K-Nearest Neighbors (KNN) and Potential Energy Method (PE). This method is formed due to the physical metaphor of potential energy and the analogy: examples of each class attract points and the 'winning' class attracts more. In this paper, I would like to work on the case study on the performance of these classifiers using the applet by E.M.Mirkes, University of Leicester [A1] and MATLAB program developed for this case study. Before formal experiments, for MATLAB program, the data pre-processing is implemented since in the database, not all data has the same scale which means it is unable to compare different data directly.

**Data Pre-processing**
The on-line databases are downloaded from UCL data website and they are saved in csv files. I sort the excel file with their specific class. Hence in the database excel file, the first column is the class label for each training example and the attributes of database starts from the second column 2. The data are then normalized with z-score normalization [1]. The z-score normalization is defined as the data subtracts the attribute mean first and divide by the attribute standard deviation. Consider a data attribute $X_{attribute} = \{X_1, X_2, \ldots, X_n\}$ if the attribute has n instances. The normalization of this attribute is computing the vector

$\widehat{X}_{attribute} = \{\widehat{X}_1, \widehat{X}_2, \ldots, \widehat{X}_n\}$ such that $\widehat{X}_i = \frac{X_i - \mu}{\sigma}$, where $\mu = \frac{1}{n}\sum_{i=1}^{n} X_i$ and

$\sigma^2 = \frac{1}{n}\sum_{i=1}^{n}(X_i - \mu)$. Normalization allows the comparison between different datasets as it eliminates the effects of gross influences. This means the comparison between different scaled datasets, in my experiment the comparison between different sets of attributes is enabled.

**K Nearest Neighbor**
The first introduced classifier is the K-Nearest Neighbor or 'KNN' for short. This classifier is memory-based and it requires no model to be fit. Consider the test point $x_{(test)}$ and the points in the training set Y where $Y = \{y_1, y_2, \ldots, y_n\}$ if the training set has n points. Then for the test point, the k training points from the training set Y are found that has closest distance to $x_{(test)}$. $x_{(test)}$ is classified as using majority vote among the k neighbors. If the features are real-valued, the Euclidean distance is used in feature space:

$$d_i = ||y_i - x_{(test)}||,$$

where i is the specific index of the points in the training set. For the computational task, [A3] is used as the algorithm for KNN. An enhanced KNN algorithm is introduced as well. It combines the classical KNN algorithm with majority voting and the idea of potential energy when selecting the local k nearest neighbor. Introducing the weight $w_i = 1/d(\mathbf{y}, \mathbf{z})^2$ for i=1,…,k and the last line of algorithm [A3] can be replaced by:

$$c_z = \text{argmax}_{v \in L} \sum_{y \in N} w_i \times I(v = class(c_y));$$

This method is much less sensitive to the choice of k. In this case, the weight factor is taken to be the reciprocal of the squared distance. In my case study, the classical KNN algorithm is used.

The classifier was first studied in 1967 by Fix and Hodges [2]. T.Cover & P.E.Hart published a famous result [3] that shows the nearest neighbor decision rule assigns to an unclassified sample point the classification of the nearest of a set of previously classified points. They also discovered that, for all distributions, the probability of error for the nearest neighbor rule is bounded above by twice the Bayes probability of error.

**Potential Energy Classifier**
The second classifier is the Potential Energy Classifier or 'PE' for short. The algorithm of PE is in the instruction for the on-line applet [A1]. Consider the training examples are labeled and all of the classes are given. Construct a function $U_i(x)$ to be the function of 'potential energy' of point x in the database for each (ith) class. Then the data point x is classified with the class which has the highest 'potential energy' $U_i(x)$. The 'potential energy' is constructed as:

$$U_i(x) = \sum_{\text{all } y_j \text{ classified with class } i} f(x - y_j),$$

where $y_j$ are all the training examples with specific class label i and the function f or named energy potential has many forms. For Yukawa potential, the function f is defined as:

$$f(x-y) = \frac{e^{-\frac{\|x-y\|}{r}}}{\|x-y\|},$$

where r is the radius of interaction and $\|x-y\|$ is the Euclidean distance. For the Gaussian potential, the function f is defined as:

$$f(x-y) = \frac{e^{-\frac{\|x-y\|^2}{r^2}}}{\|x-y\|},$$

where $\|x-y\|$ uses the Euclidian distance and r is the radius of interaction.

The parameter named 'Effective radius of interaction' is the radius of interaction r in the calculation of energy potentials. The choice of r is different for different experiments. For the on-line applet, the radius is chosen between 1 and 200 since the height of the work desk is 400.

For experiments using real life benchmark database, I first computed the average distance for each database matrix and then calculate the square root of this average distance. The interactive radius used here is the specific percentage of the square root average distance. (i.e. 10%,20%,…,200%) The Leave-One-Out Cross-Validation is applied for potential energy with different values of interactive radius. The result is in [Graph 1.1] for PE Y and [Graph 1.2] for PE G. From these graphs, it is easy to see for the radius percentage of 10%, all databases seems to have relatively small error rate. Hence for the experiment in Chapter 3, the 10% square root average distance is used as the parameter for potential energy classifier.

Radial Basis Function
The function of energy potential for Yukawa and Gaussian are different forms of radial basis functions. Some general definitions and properties of radial basis function are described in [4]. Radial basis function has several applications such as collocation methods and the combined methods for elliptic boundary value problems. [5] The Yukawa and Gaussian potentials are used as the energy potential functions in PE classifier. These functions are plotted against the (x-y) with r=10 for easy view in [Graph 1.3] and [Graph 1.4]. From this graph, both potential functions has singularity when x-y=0. The functions reach positive infinity when x-y→0 and decay as (x-y) increases. This is a good way to separate the data points that are relative far away from the test points and has less effect on the classification result.

**Condensed Nearest Neighbor Rule (Enhanced Hart Algorithm)**
In 1967, Hart introduced the condensed nearest neighbor decision rule or short for CNN and the Hart Algorithm in [6]. In [7], N. Bhatia and Vandana compared many nearest neighbor techniques including CNN. They applied CNN on the training set to minimize the time complexity without losing relevant information. For application of KNN classifier, the CNN rule is applied on the training set to reduce the size of training set. Then for KNN classifier, the test points are the same as one without applying CNN on the database. The CNN rule uses the Hart Algorithm to determine a consistent subset of the original sample set but it will not find a minimal consistent subset, the algorithm used for on-line algorithm and MATLAB program is introduced later in this chapter.

The applet of KNN uses CNN rule to reduce the training set for KNN classification. The size of the database is finite. If considering the database to be a finite set, then the rule is applied to find the subset U of this database. One comment is that if applying 1NN rule on prototypes set U, the result is 'almost' as successful as the result of applying 1NN on initial database. Hence for many experiments using KNN or CNN, the value k=1 is chosen.

The application of the Hart algorithm is, consider set X with n data to be the set of training examples. Reorder the set X with some techniques and numerate the ordered set:
$$X = \{x_1, x_2, \ldots, x_n\}.$$
Starting with put $x_1$ in to the set U, i,e, $U = \{x_1\}$. Let element in the set U be called 'prototype'. Then the rule scans all elements of X. During the scanning, if the nearest prototype of the example x from U does not have the same label as x, this example will be put inside the set U. The scanning process will carry on if there are more prototypes are put in U and it will stop when there are no more prototypes added. The new set U is used as the database for classification. One remark is that: during the scanning process, if x is included in U as a prototype then it is excluded from X and does not be tested in the following scans. The CNN rule is applied to the training set of the database. Then the KNN classifier is applied using this training set and the test points are still from the original database. The classification result applied 1CNN (CNN with KNN has k=1) will have almost the same error rate as 1NN (KNN with k=1).

**Advanced procedure for ordering**
For computation of CNN rule, a special technique of ordering is used. Consider X to be the set of examples. Let x be a labeled example (with specific class label). The external examples for x are the data with different labels. Then, find the external example y of x which is closest to x. The other example x' is found as the closest data to y but with the same label as x. The distance $\|x' - y\|$ is never exceed the distance $\|x - y\|$. This is because if you think the data y as the centre of a circle with radius $\|x - y\|$, the closest data x' with the same label of x always in this circle hence the ratio $a(x) = \|x' - y\| / \|x - y\|$ lies in the interval [0,1]. The ratio a(x) is called the border ratio. The next step is to order all the examples in X accordance with the values of a(x), in the descending order. This ordering gives preferences to the borders of the classes for inclusion in the set of prototypes U.

The error analysis of the classifiers is based on the Cross-validation for KNN, PE with Yukawa energy potential and PE with Gaussian energy potential. For CNN, Cross-Validation can only gives errors since the database after CNN is a whole new database, which completely changed into a new classification database. The CV result will be absolutely useless.

**Leave-One-Out Cross-Validation**
Cross-validation is one of the most widely used methods for estimating prediction error. In this method, it directly estimates the expected extra-sample error :
$$\text{Error} = E[L(Y, \hat{f}(X))],$$
which is the average generalization error when the method $\hat{f}(X)$ is applied to an independent test sample from the joint distribution of X and Y. In [8], R. Kohavi explains some of the assumptions made by different estimation methods including Cross-Validation and represent examples where they could fail.

The Leave-One-Out Cross-Validation is a special case of K-Fold Cross-Validation. In Leave-One-Out Cross-Validation, each point in the training set is considered as the test point. When a test point exists, it is excluded from the training set and use the reduced training set is used as the training set for Cross-Validation.

To compare the classification results between two classifiers, correlation coefficient could be used to test the relation of them. The famous measure is the Pearson Product-moment correlation coefficient.

**Pearson Product-moment correlation coefficient**
The Pearson product-moment correlation coefficient between two variables is defined as the covariance of the two variables divided by the product of their standard deviations. Consider X and Y are random variables. The correlation coefficient $\rho_{X,Y}$ is defined as:
$$\rho_{X,Y} = \frac{\text{cov}(X,Y)}{\sigma_X \sigma_Y} = \frac{E[(X-\mu_X)(Y-\mu_Y)]}{\sigma_X \sigma_Y}.$$
Where $\mu_X$ and $\mu_Y$ are means of random variables X and Y, $\sigma_X$ and $\sigma_Y$ are the standard deviation of random variables X and Y. The absolute value of the correlation coefficient is less than or equal to 1. Correlations equal to 1 or -1 correspond to data points lying exactly on a line. Hence, coefficients that close to 1 means strong linear relation, close to 0 means independent relation and close to -1 means negative linear relation.

To observe the rate of change between different algorithms, information gain could be computed. The introduction on information gain starts with introducing the entropy.

**Entropy**
The definitions of entropy, conditional entropy and information gain are from [9]. Entropy is a measure of uncertainty of a random variable. Let X be a discrete random variable with alphabet $\chi$ and probability mass function $p(x) = \Pr\{X = x\}, x \in \chi$. The entropy $H(X)$ of a discrete random variable X is defined by
$$H(X) = -\sum_{x \in \chi} p(x) \log p(x).$$
This log is to the base 2.

**Conditional Entropy**
Let $(X, Y)$ be a single vector-valued random variable. Consider $p(x, y)$ to be the joint distribution. If $(X, Y) \sim p(x, y)$, the conditional entropy $H(Y|X)$ is defined as:
$$H(Y|X) = \sum_{x \in \chi} p(x) H(Y|X = x) = -\sum_{x \in \chi} \sum_{y \in \mathcal{Y}} p(x, y) \log p(y|x).$$

**Information Gain**
The information gain of Y given X is then defined as:
$$IG(Y|X) = H(Y) - H(Y|X).$$
The information gain is the reduction in the uncertainty of Y due to the knowledge of X.

The hypothesis test could also be applied to check the performance of different classifiers. In [10], the definition of McNemar's test is introduced.

**McNemar's Test**

Two algorithms are used to train two classifiers on the training and set and test them on the validation set and compute their errors if a training set and a validation set are given. The contingency table of errors is constructed first:

| $e_{00}$: Number of examples misclassified by both | $e_{01}$: Number of examples misclassified by 1 but not 2 |
|---|---|
| $e_{10}$: Number of examples misclassified by 2 but not 1 | $e_{11}$: Number of examples correctly classified by both |

Under the null hypothesis that the classification algorithms have the same error rate, $e_{01} = e_{10}$ is expected and these are equal to $(e_{01} + e_{10})/2$. The chi-square statistic with one degree of freedom is:

$$\frac{(|e_{01} - e_{10}| - 1)^2}{e_{01} + e_{10}} \sim \mathcal{X}_1^2$$

McNemar's test rejects the hypothesis that the two classification algorithms have the same error rate at significance level $\alpha$ if this value is greater than $\mathcal{X}_{\alpha,1}^2$.

## 2. Analysis using on-line applet

The first part of analysis is using the Applet. The applet generates the imaginary data (each coloured point) and using different methods (KNN and PE) to draw classification maps, CNN reduction for KNN method, leave-one-out cross validation for each method or to test a specific point in a 2-d plane. The full instruction for every functional panel is explained on the website. By using this on-line applet, the performances of the two methods are tested. For the radius of PE classifier in the applet, it is chosen between 1 and 200.

A database consists 140 data with two classes (blue and red) are constructed. The formation of this database is:
1. For each class, generate 3 sets of 20 scatter data points next to each other. For different class, the data points are almost independent of each other.
2. For each class, generate 10 random data points in the screen.

Applying the Leave-One-Out Cross-Validation for KNN, the error ratio is plot as a function of k, where k is an odd number and k=1, 5, 11, 15, 21, 25, 31 and 35. The result is in [Graph 2.1]. The error ratios are bounded between [0.12, 0.16] with the minimum error ratio occur at k=11. It generally decreases after k=1 and has an approximate horizontal trend from k=5 to k=35.

Applying the Leave-One-Out Cross-Validation for PE with different potential, the error ratio is plot as a function of r, where r is an integer value and r=10, 20, 30, 40, 50, 60, 70, and 80. The result is in [Graph 2.2]. For PE method with Yukawa potential (PEY in the graph), the error ratios are bounded between [0.12, 0.14], with the minimum error at r=30. For PE method with Gaussian potential (PEG in the graph), the error ratios are bounded between [0.12, 0.16], with the minimum error at r=40 to 60. The maximum change of error ratios for PEG is a bit higher than the change of error ratios for PEY but they all converged to a ratio of around 0.135 with almost same speed of convergence. The error analysis shows some conclusion that there might be similarity in the performance between KNN methods and PE methods as in general, the two graphs of error analysis are similar.

To test if there is similarity in the results using different methods, the indicator I used is the Pearson's correlation coefficient. Since for this graph it is difficult to judge by eye, I construct an algorithm for MATLAB to edit the graph, before applying this indicator.

**MA.1 MATLAB algorithm for editing graph**

Input: The png file of graph.
1. Read the png file of graph and saved as a three dimensional int8 matrix
2. Rescale the png file if it is out of size.
3. Scanning the matrix and assign the different RGB scale point to the RGB coordinate of either Blue or Red color. The method of change in the coordinate is simply transferring it to the nearby Blue or Red coordinate.
4. Transform the result matrix into image

Output: The image of classification map that only has blue or red color. (Neither any data point nor any other images exists on the map)

The result classification map is used in the calculation for correlation coefficients. When the image is read into MATLAB, it is saved as an $m \times n \times 3$ matrix. Consider the calculation for correlation coefficient between graph 1 and graph 2. Let X be the random variable for the specific data point in graph 1 and Y be the random variable for the data point in graph 2. Both X and Y has two outcomes, they are defined as:

$$X_i = \begin{cases} 1 & \text{if the data point is red} \\ -1 & \text{if the data point is blue} \end{cases}, Y_i = \begin{cases} 1 & \text{if the data point is red} \\ -1 & \text{if the data point is blue} \end{cases}$$

where $X_i, Y_i$ represents the specific data point in the graph where $i = 1, \ldots, m \times n$.

The variances of X and Y are computed using the equation:
$$\text{Var}(X) = E(X^2) - (E(X))^2$$
where $E(X) = \sum_{i=1}^{k} X_i \Pr(X = X_i)$ and $k = m \times n$ which is the first two dimensions of the matrix. The computation of covariance of X and Y is more complicated. Let X and Y be the random variable of the colour for specific data point on graph 1 and 2. Hence the event (X,Y) is defined as:

$$(X, Y) = \begin{cases} (1, 1) & \text{if the data point is red on both graphs} \\ (1, -1) & \text{if the data point is red on graph1 and blue on graph 2} \\ (-1, 1) & \text{if the data point is blue on graph1 and red on graph 2} \\ (-1, -1) & \text{if the data point is blue on both graphs} \end{cases}$$

The covariance of X and Y is computed as:
$$\text{Cov}(X, Y) = E(XY) - E(X)E(Y)$$

where $E(XY) = \sum_{i=1}^{k} \sum_{j=1}^{k} X_i Y_j \Pr(X = X_i, Y = Y_j)$.

The correlation coefficient is then computed using this covariance divided by the product of the marginal standard deviation. Since each classification map is the result of performance for different methods, the correlation coefficients between two classification maps could measure the similarity for the result. If the coefficients are close to 1, it shows very high similarity in the results using two methods in classification problems; hence it means the performances of the methods are very similar. If the coefficients are close to 0, it shows there is no similarity in the performance and if the coefficients are close to -1, it shows there is negative correlation relation between two methods.

The correlation coefficients between two classification maps generated from the applet for all possible combinations are calculated. The result is in [Graph 2.3]. The correlation coefficients between 1NN and PE with both potentials are decreasing as the number of r increases. The change in the correlation coefficients for PE with Yukawa potential is more than the change in the correlation coefficients for PE with Gaussian potential. The correlation coefficients between 1CNN and PE are generally increasing. Except for the coefficients between 1NN and PEY, the rest of coefficients converge to approximated 0.78. From the result, it can conclude that the classification results between KNN and PE are more similar before applying CNN reduction rule.

Another phenomenon is found that, the density of the data could affect on the classification results. Generate three new sets of databases with 2 classes, blue and red. For each set of data, there are 20 red data points. The numbers of blue points increase from set 1 to set 3, the numbers are: 20, 100 and 200. The classification maps of the three sets are in [Graph 2.4]. No random data points are generated. Applying KNN, CNN methods with k=1 and applying PE Y and PE G with r=30. I calculate the correlation coefficients and plot the bar chart. [Graph 2.5]

From the bar chart, the correlation coefficients in set 1 are almost the same. From set 2, the coefficients for PE Y start to decrease and reach the minimum value for set 3. These show the results of the classification problems are less similar. The decrease in the similarity means when the 'weight' of data is getting higher, the classification results applying PE method with Yukawa potential would be affected. Here 'weight' means the number of data points in specific class. The Leave-One-Out Cross-Validation errors also exist when the density of the data points with specific class are relatively high compare to the data points to the rest of classes. However, if normalized potential function is applied rather than the sum of total energy potential, this error will vanish.

From the two experiments using applet, it could be conclude that the performances of the classification between KNN and PE method with both potentials are similar to the performances of classification between CNN and PE method.

3. **Analysis using Real Life Benchmark Data**

In this section, 6 databases of real life benchmark are used for analysis. The 5 databases are collected from UCL data website and the address is given in [Website 3.1] with descriptions of databases. The Party database is used in [3]. The Party dataset represents some of the results of presidential elections in the USA over the period from 1860 until 1992. There are 12 attributes for this database with either yes 'y' or no 'n' to the specific question. There are two classes, P-class corresponds to the elections with P-party candidate won; O-class corresponds to the elections with O-party candidate won. Some basic descriptions of the 6 databases are in [Table 3.1]. The algorithm of the MATLAB program is simply computed following the instructions on the applet website. After applying KNN, CNN and Potential energy method on the 6 real databases, I used correlation coefficients, information gain and McNemar's test as indicator to see the performance for different classification methods on 6 real databases. For the radius of PE classifier, it is the radius of the normalized data.

Before calculating the correlation coefficients, I used Leave-One-Out Cross-validation (LOO CV) on classification methods KNN, PE with Yukawa potential and PE with Gaussian potential. The parameter of KNN is 1 and the parameter of PE classifier with both energy potentials is chosen as 10% of square root average distance. The number of errors is the number of misclassifications using LOO CV, the error ratio is defined as the number of errors divide the total number of instances. The error ratio for CNN is the ratio of outliers' number and total instances number. The error percentages for different databases with different classification methods are in [Graph 3.1]. From the table, it shows that the error percentages are bounded between [0, 0.4]. Iris database has the smallest error ratio for all 4 classification methods. Except the Transfusion database, the error ratio of 1NN and 1CNN are almost the same for the rest five databases. For database Haberman and Party, the error ratios are a bit higher than the rest methods. Hence it could conclude that, the performance of the four classifiers for these two databases is not very good compared to other methods. Except for the Transfusion databases, the error ratio behaves quite similar with 'U' shaped distribution of the histogram. But the difference is not significant high. Hence it could conclude that for these three databases, these five methods behave similarly. The error ratio is relatively high for Transfusion database. And it has the highest error ratio for CNN method.

Before computing the correlation coefficient, I would like to construct a MATLAB program to transform the results of strings into natural numbers. The output from MATLAB program is constructed as an array of strings. Then the result vector of numbers is used in the calculation of correlation coefficients. The coefficients are between -1 and 1 with 1 represents that two methods have very the same classification results, 0 represents that two methods have no similar classification results and -1 represents that two methods have negative relation for the classification results. Note for the classification result of CNN method, the correlation coefficients are calculated between result of CNN and a corresponding subset of the original database with labels. If not, it will have errors since two vectors have different lengths.

The correlation coefficients between different methods for 6 databases are calculated and the result is in [Graph 3.2]. In this table, the correlation coefficient of each combination is calculated. In general, all databases except for Transfusion database have relatively high coefficients and the coefficients between different pairs of classifiers are very similar. Hence it shows that the behaviors of four classifiers are almost the same. For Transfusion database, the coefficients between different pairs of classifiers are different. The coefficients between 1CNN and other classifiers are relatively higher than the coefficients between 1NN and other classifiers. It may shows that, applying CNN rule on this database may give a more similar classification results to potential energy methods. The coefficient between PE Y and PE G behaves just like the one from other database. The difference in the result may because of the choice of the parameter for k in KNN and r in potential energy method.

To observe the change in the classification results between two different methods, I calculated the information gain. Following the definition of information gain in the introduction, using the computation algorithm from [A2], the information gain could be computed in the following way.

Consider an vector column $X$ and another vector column $Y$, they are symbolic attributes or specifically, they are labels of classification results. Let $X$ have arity $A_X$ and $Y$ have arity $A_Y$. Let $X_k \in \{0,1,\ldots,A_X - 1\}$ be the value of kth element of column X and let $Y_k \in \{0,1,\ldots,A_Y - 1\}$ be the value of kth element of column Y. Assume both columns have length N hence $k \in \{0,1,\ldots,N-1\}$. The information gain of Y given X is defined as:

$$\text{InfoGain}(X,Y) = \text{InfoGain}(Y|X) = \left(\sum_{j=0}^{A_Y-1} f\left(\frac{V(j)}{N}\right)\right) - \frac{1}{N}\sum_{i=0}^{A_X-1}\left(U(i)\sum_{j=0}^{A_Y-1} f\left(\frac{W(i,j)}{U(i)}\right)\right),$$

where $W(i,j) = $ Number of $(X_k, Y_k)$ pairs in which $X_k = i$ and $Y_k = j, i \in \{0,1,\ldots,A_X - 1\}$ and $j \in \{0,1,\ldots,A_Y - 1\}$. And vectors U and V are defined as:

$$U(i) = \sum_{j=0}^{A_Y-1} W(i,j), \ V(j) = \sum_{i=0}^{A_X-1} W(i,j), N = \sum_i \sum_j W(i,j).$$

The computed information gain between each method is saved in [Graph 3.3]. From this bar chart, the Glass database has the highest information gains and the information gains of Haberman, Ionosphere, Party and Transfusion databases are relatively low. For specific database, all six information gains are similar. Hence it may conclude that for the 6 data bases, the updates in information between all pairs of classifiers are quite similar.

I also compute the information gain of the original class label given classification results. The result is saved in [Graph 3.4]. From this bar chart, the Glass database has the highest information gain in the six databases, and the Haberman, Ionosphere, Party and Transfusion databases have relatively the lower information gains. For each specific database, the four information gains are very similar. Hence it may conclude that, for each database, the update of original data label given the classification results are the same and the change in the classification results are the same for the four different classifiers.

To compare the performance between two algorithms, I used McNemar's test. Consider the errors are the number of misclassification of Leave-One-Out Cross-Validation for specific method. Compute the contingency error table and use the test statistics compare with $\mathcal{X}^2_{0.05,1}$. The hypothesis test is constructed as:
$H_0$: Two classification algorithms have the same error rate.
$H_1$: Two classification algorithms have the different error rate.
The test statistic at 5% significant level has a value of $\mathcal{X}^2_{0.05,1}$=3.84. The result of the test is saved in [Table 3.2]. From the table, it shows that except for Transfusion database, almost all of the test results show that different pairs of classifiers has the same error rate. For Transfusion database, the only test result gives the same error rate is the one between CNN and PE G.

From the two experiments above, the results for Transfusion database are different compare to the rest five databases. Hence for this database, I compare the results using previous experiment for different interaction radius. The error analysis for PE method with different percentages of the square root average distance is computed. (i.e. for p=10%, 20%, …, 200%) The result is in [Graph 3.5]. It shows that the error of PE G decreases from p=10% and converge to approximately the same error ratio of PE Y after p=30%. The error of PE Y does not change a lot. The correlation coefficients between different classifiers are computed against p=10%,20%, …, 200%. The result is in [Table 3.3] and the bar chart is in [Graph 3.6]. The coefficients are relatively high for p=10% and they decrease when p is larger and some of them have negative values. Hence for result of coefficients, using p=10% is a better option.

The information gain of Original Label given different classifiers for different radius is also computed. The result is in [Graph 3.7]. In this graph, the information gains are highest when p=10%. It also indicates that using the interaction radius with percentage 10% of the square root average distance is a better option. The McNemar's test is computed and the result is in [Table 3.4]. The test results are almost the same hence all percentages may be reasonable. Therefore in general, choosing p=10% is a good parameter for Transfusion database.

## 4. Conclusions

By using applet, the analysis result shows that for some specific database, the classification result after implementing CNN is more similar to the classification result using PE method than the result using KNN. PE method with Yukawa potential is very sensitive to the density of database. The higher density of data points in a specific area, the worse the classification result it will be. But the lack of the result could be eliminated by the normalized energy potential. PE method in general may have similar behavior to KNN and CNN. For the 6 real life benchmark databases, it could conclude that the four classifiers behave similarly. Except for Transfusion database, by using McNemar's test, the error rates for the rest databases between two classifiers in each combination are the same. The future research could concentrate on the error analysis and optimization in choice of parameters for potential energy classifier.

# Appendix

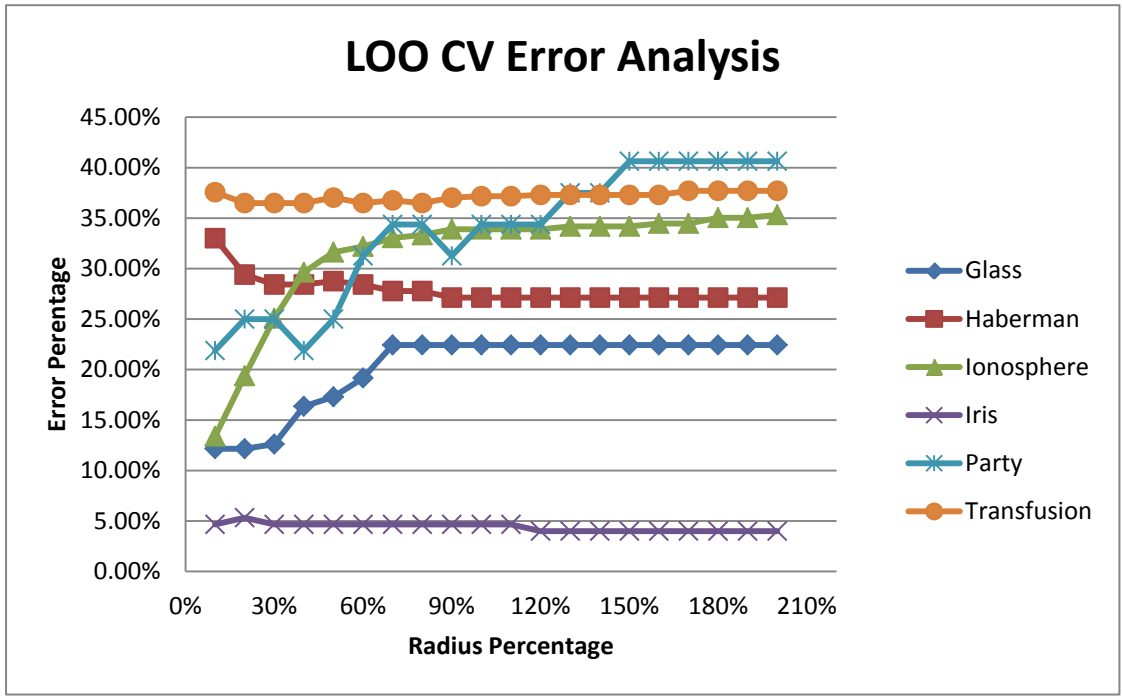

[Graph 1.1] The graph of Leave-One-Out Cross-Validation error analysis of Potential Energy Classifier with Yukawa Potential for 6 real life benchmark databases

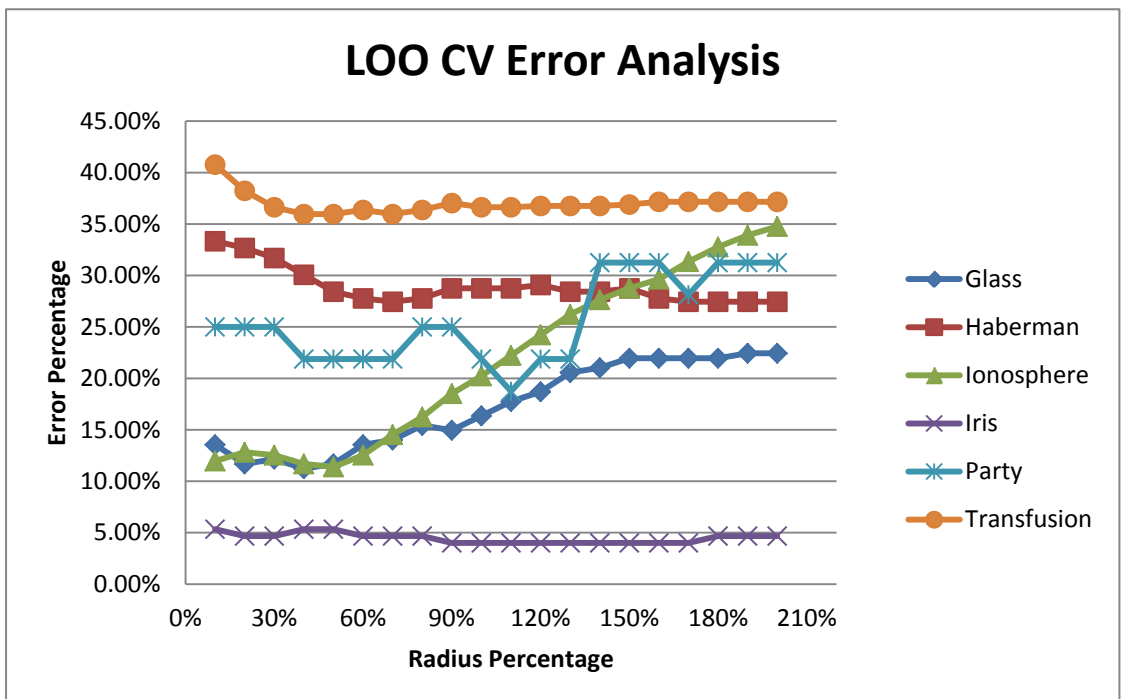

[Graph 1.2] The graph of Leave-One-Out Cross-Validation error analysis of Potential Energy Classifier with Gaussian Potential for 6 real life benchmark databases

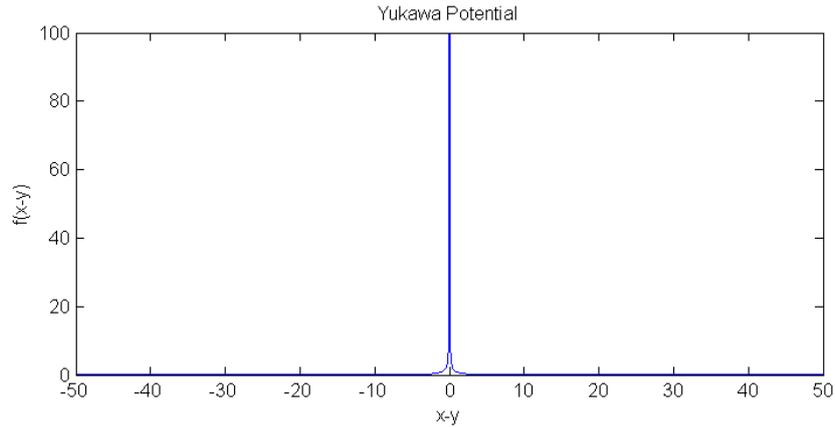

[Graph 1.3] The graph of Yukawa potential function f(x-y) against x-y. The range of x-y is from -50 to 50 with 0.01 between each point of x-y.

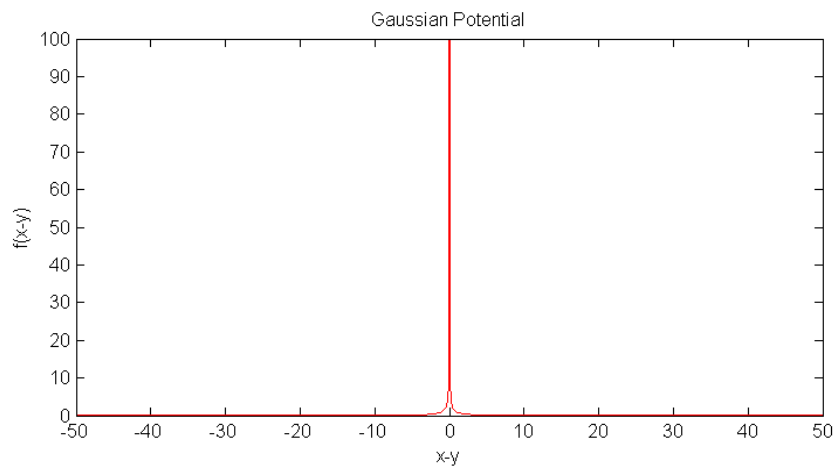

[Graph 1.4] The graph of Gaussian potential function f(x-y) against x-y. The range of x-y is from -50 to 50 with 0.01 between each point of x-y.

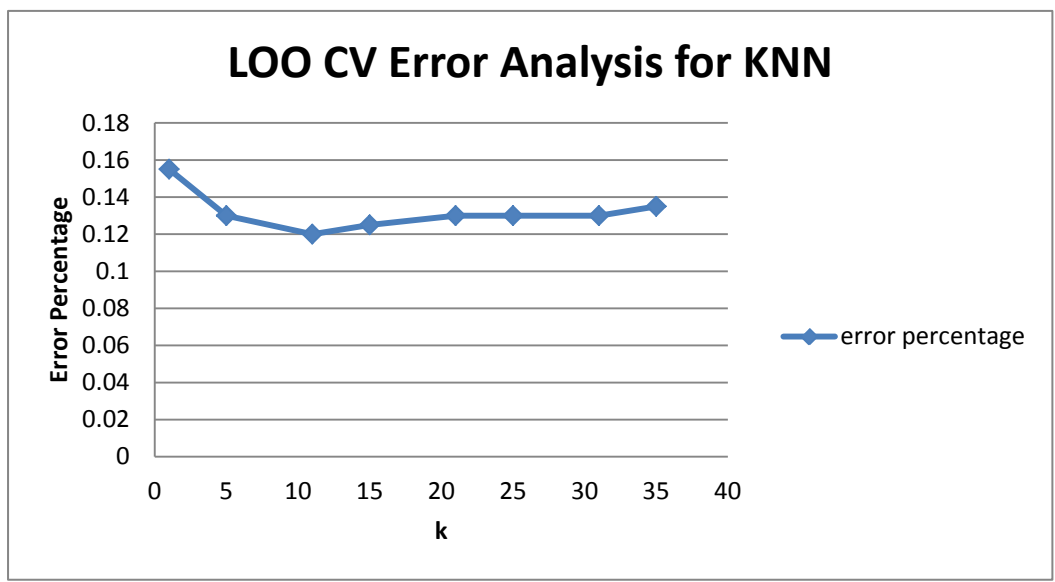

[Graph 2.1] The graph of Leave-One-Out Cross-Validation error analysis for KNN method using applet for k=1,5,11,15,…,35

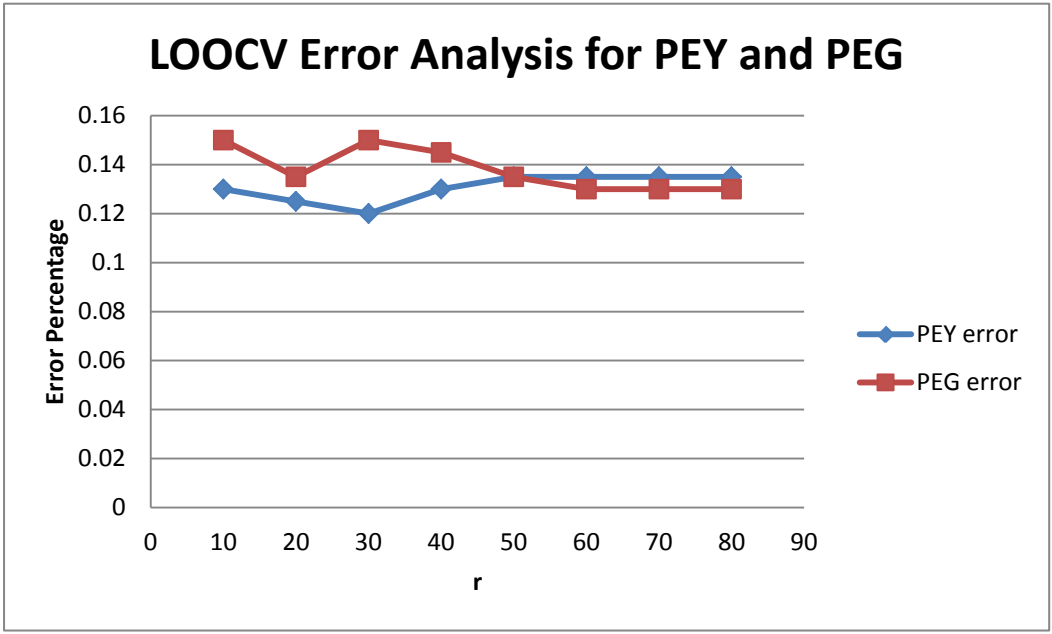

[Graph 2.2] The graph of Leave-One-Out Cross-Validation error analysis for PE method using applet with specific energy potential. The radius of interaction r=10,20,…,80 and the maximum of this value is 200.

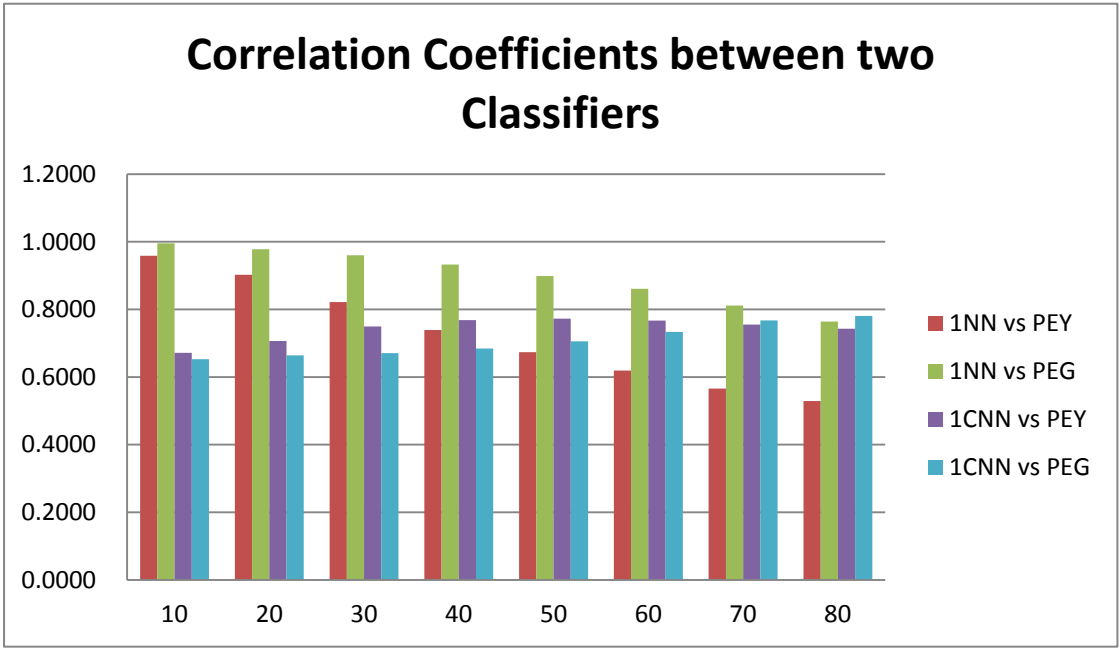

[Graph 2.3] Bar Chart of correlation coefficients between two classifiers for each possible combination using applet

**Set1**

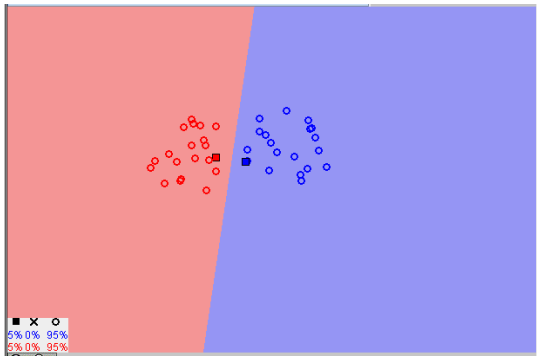

**(a)**

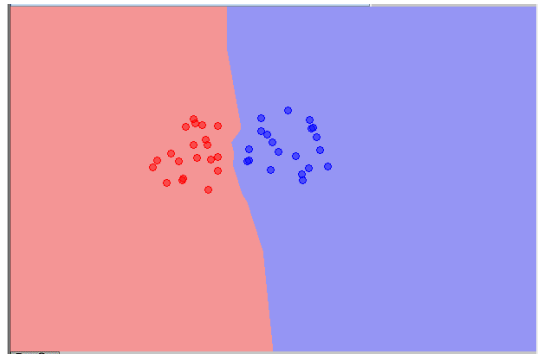

**(b)**

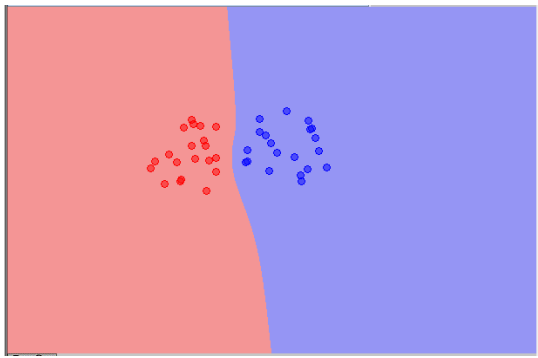

**(c)**

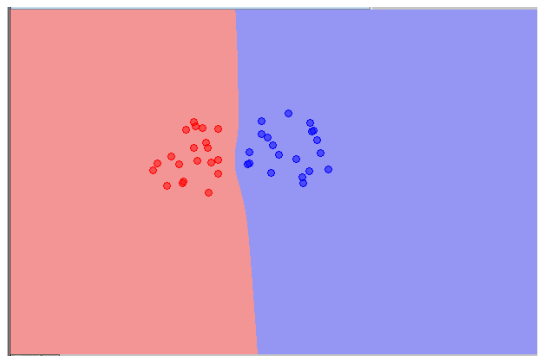

**(d)**

**Set 2**

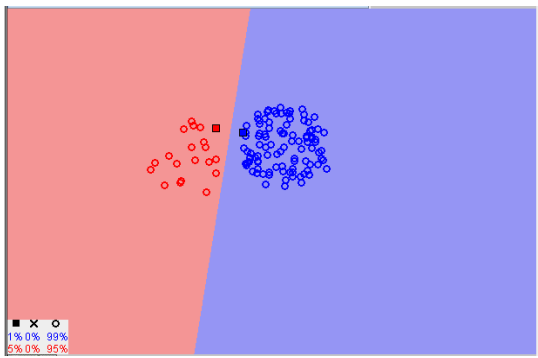

**(a)**

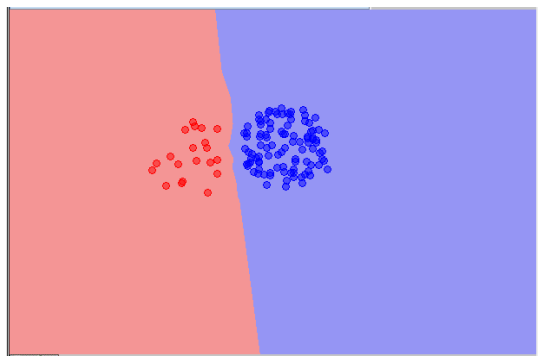

**(b)**

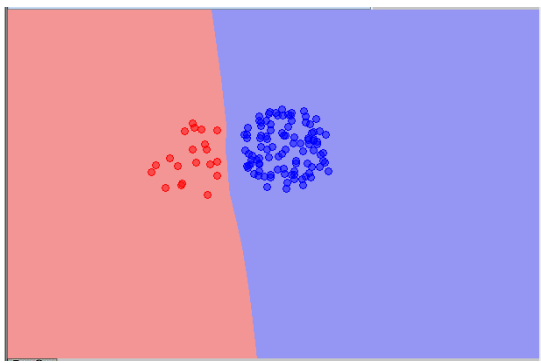

**(c)**

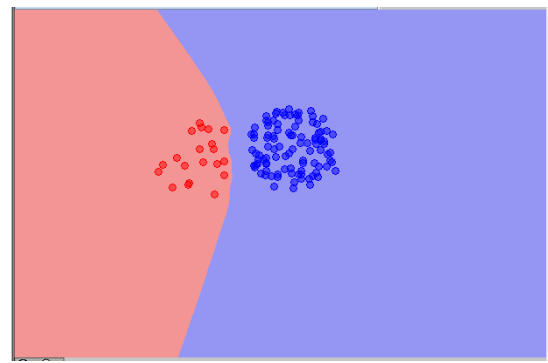

**(d)**

**Set 3**

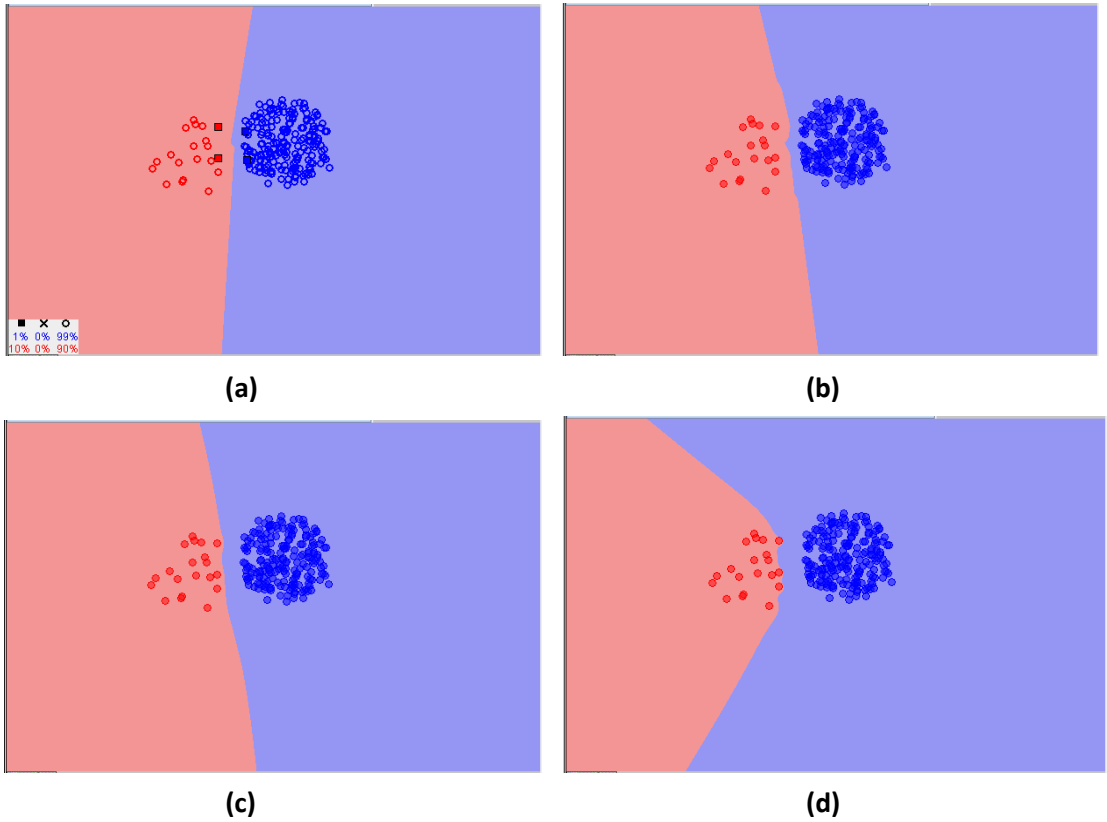

[Graph 2.4] Classification maps of three sets. For the experiment using each set of data points, different numbers of data points are used. For **Set 1**, there are 20 red data points and 20 blue data points. For **Set 2**, there are 20 red data points and 100 blue data points. For **Set 3**, there are 20 red data points and 200 blue data points. For each set of data points, there are four classification maps. **(a)** Classification map for 1CNN **(b)** Classification map for 1NN **(c)** Classification map for PE classifier with Gaussian potential using r=30 **(d)** Classification map for PE classifier with Yukawa potential using r=30

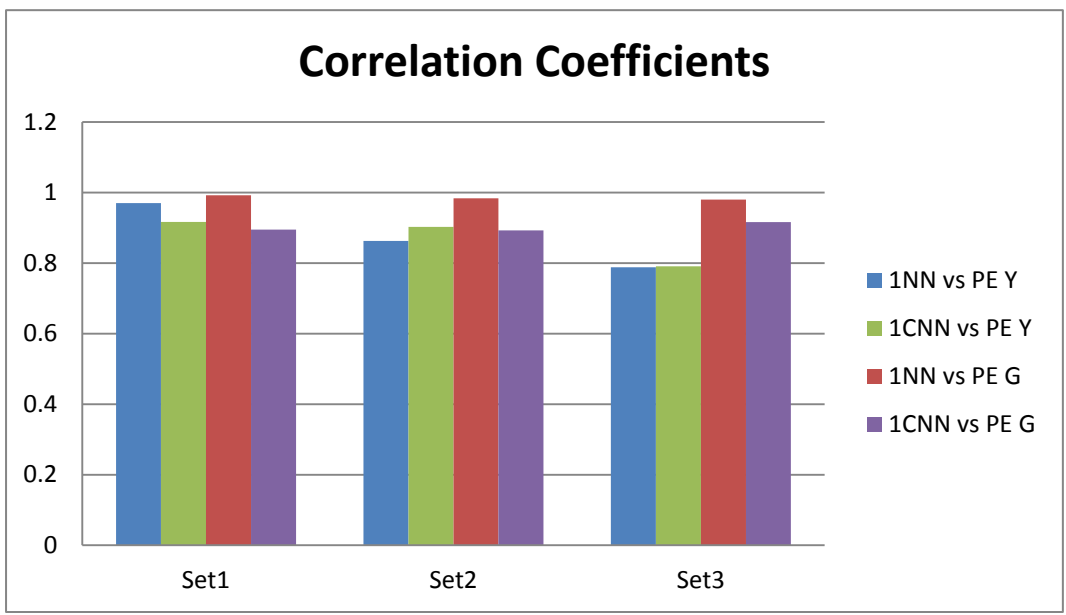

[Graph 2.5] Bar Chart of correlation coefficients between two classifiers for each combination of three sets of data points

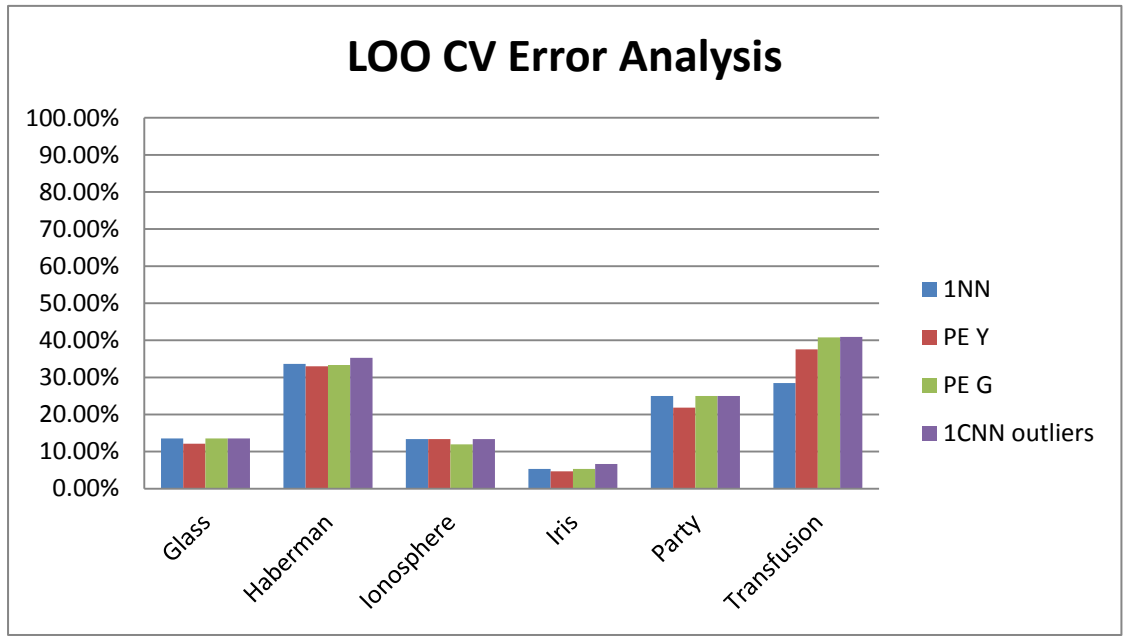

[Graph 3.1] Graph of Leave-One-Out Cross-Validation and 1CNN outliers error analysis for 6 real life benchmark databases

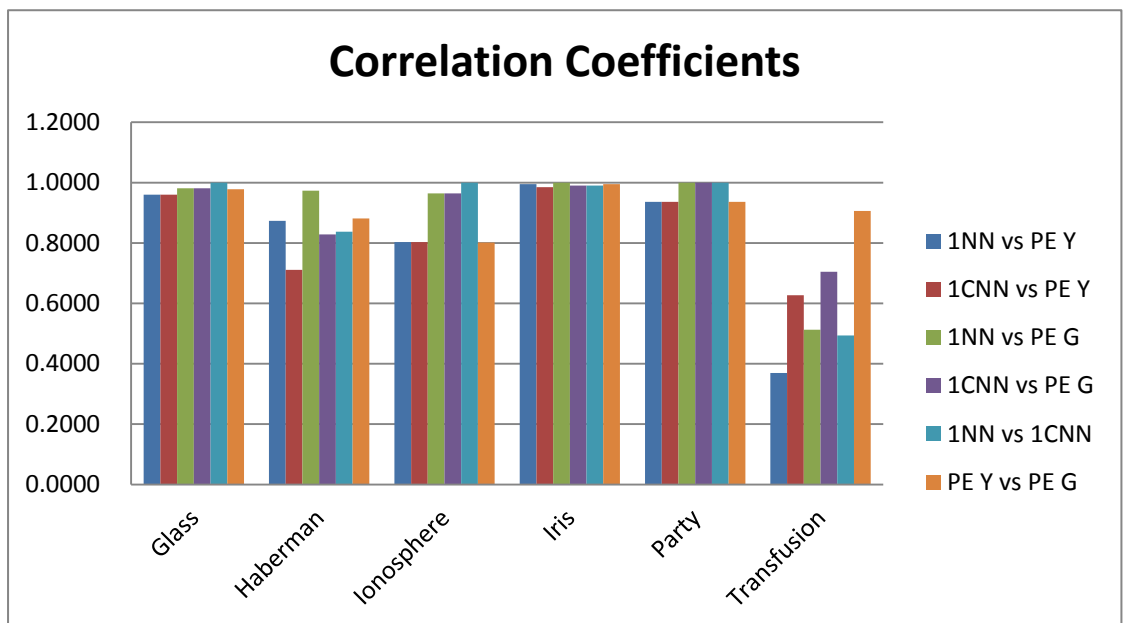

[Graph 3.2] Graph of correlation coefficients of classification results between two classifiers of each combination for 6 real life benchmark databases.

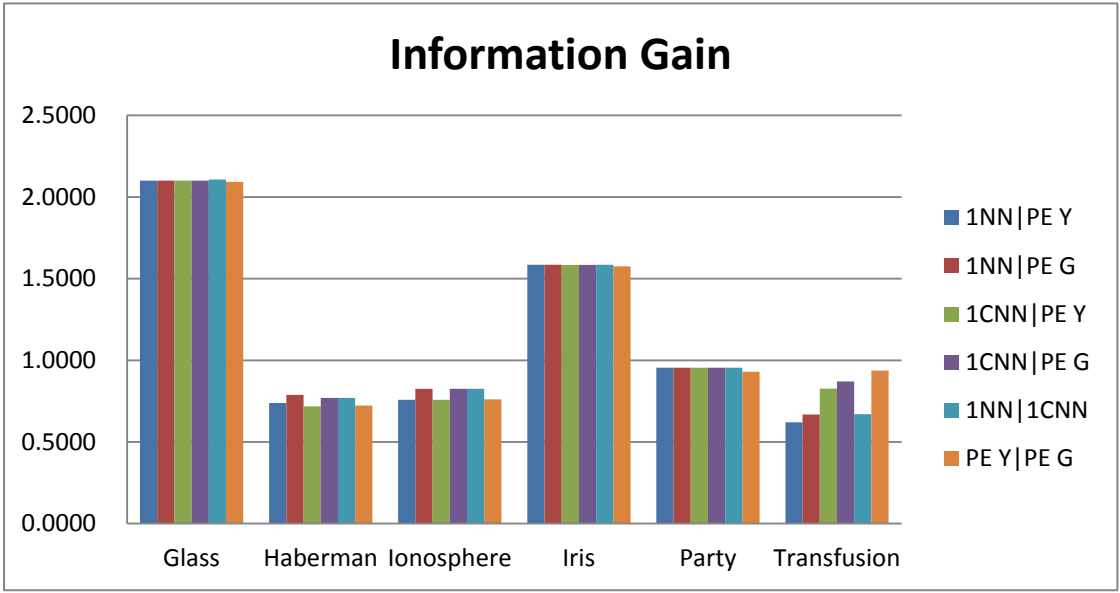

[Graph 3.3] Graph of information gain between two classification results of two classifiers for 6 real life benchmark databases.

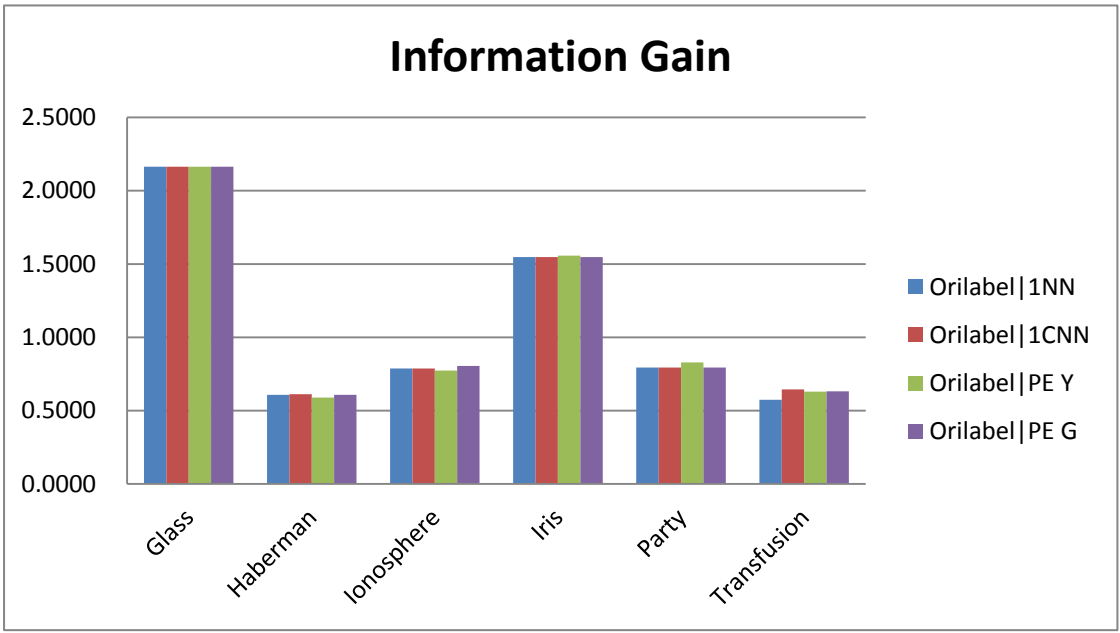

[Graph 3.4] Graph of information gain between original label and classification results of four classifiers for 6 real life benchmark databases.

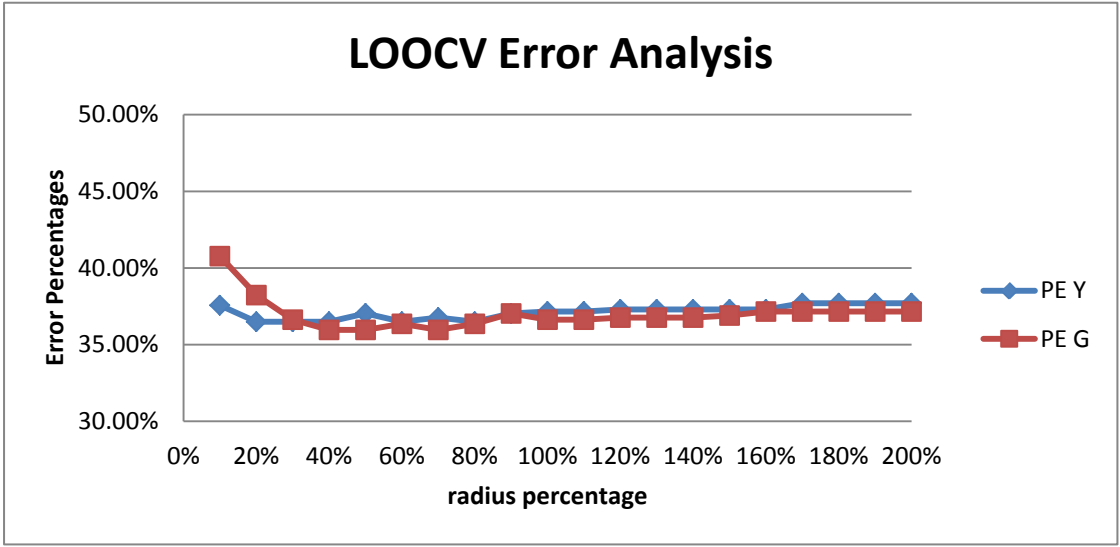

[Graph 3.5] Graph of Leave-One-Out Cross-Validation error analysis of PE classifier with Yukawa and Gaussian potentials for Transfusion Database. The radius of interaction r is taken as proportion ratio of the average distance between test points and training points. The ratio is taken from 10% to 200%.

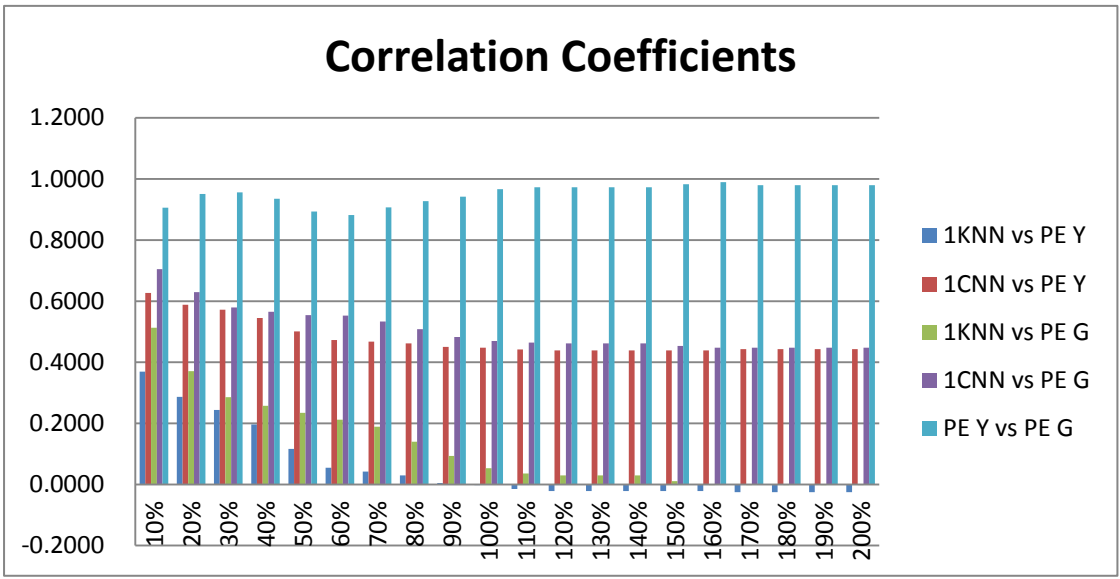

[Graph 3.6] Graph of correlation coefficients between two classification results of two classifiers of each combination for Transfusion database. The radius of interaction r is taken as proportion ratio of the average distance between test points and training points. The ratio is taken from 10% to 200%.

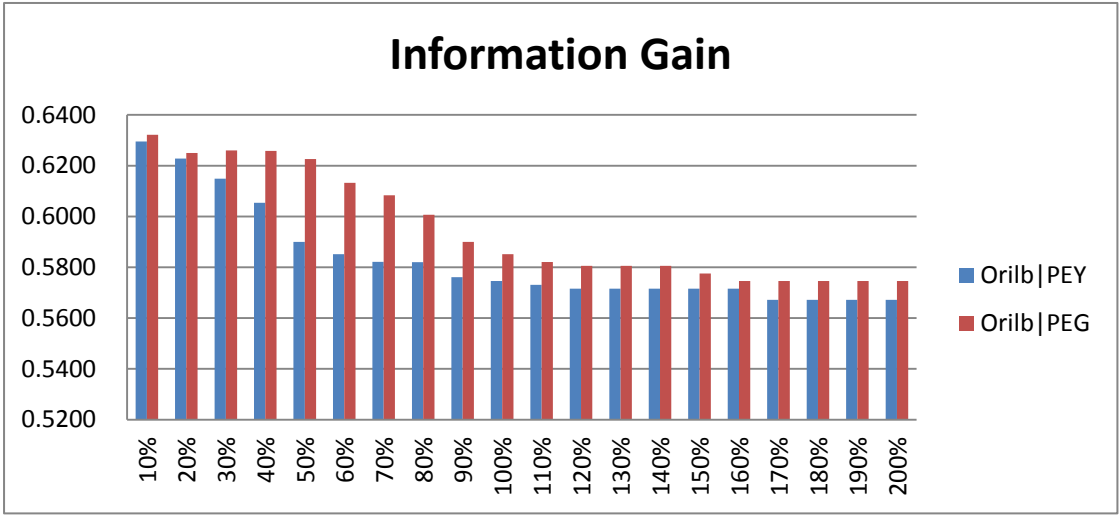

[Graph 3.7] Graph of information gain between original label and classification results of PE classifier with Yukawa and Gaussian potentials. The radius of interaction r is taken as proportion ratio of the average distance between test points and training points. The ratio is taken from 10% to 200%.

[Table 3.1] Table of Descriptions for 6 Real Life Benchmark Databases

| Name of Dataset | Instances No. | Attributes No | Class No | Attribute Characteristics |
|---|---|---|---|---|
| Glass | 214 | 10 | 7 | Real |
| Haberman | 306 | 3 | 2 | Integer |
| Ionosphere | 351 | 34 | 2 | Integer,Real |
| Iris | 150 | 4 | 3 | Real |
| Party | 384 | 12 | 2 | Categorical |
| Transfusion | 748 | 5 | 2 | Real |

[Table 3.2] Table of results using McNemar's test with '0' means the null hypothesis is accepted (two classifiers have the same error rate) and '1' means the alternative hypothesis is accepted (two classifiers have different error rates).

|  | Glass | Haberman | Ionosphere | Iris | Party | Transfusion |
|---|---|---|---|---|---|---|
| KNN vs PE Y | 0 | 0 | 0 | 0 | 0 | 1 |
| KNN vs PE G | 0 | 0 | 0 | 1 | 1 | 1 |
| CNN vs PE Y | 0 | 0 | 0 | 0 | 0 | 1 |
| CNN vs PE G | 0 | 0 | 0 | 0 | 1 | 0 |
| KNN vs CNN | 1 | 0 | 1 | 0 | 1 | 1 |
| PE Y vs PE G | 0 | 0 | 0 | 0 | 0 | 1 |

[Table 3.3] Table of Classification coefficients between classification results of two classifiers with for Transfusion Database. The radius of interaction r is taken as proportion ratio of the average distance between test points and training points. The ratio is taken from 10% to 200%.

|  | 10% | 20% | 30% | 40% | 50% | 60% | 70% |
|---|---|---|---|---|---|---|---|
| 1KNN vs PE Y | 0.3694 | 0.2865 | 0.2440 | 0.1965 | 0.1162 | 0.0548 | 0.0423 |
| 1CNN vs PE Y | 0.6269 | 0.5884 | 0.5723 | 0.5451 | 0.5009 | 0.4729 | 0.4673 |
| 1KNN vs PE G | 0.5129 | 0.3710 | 0.2858 | 0.2574 | 0.2344 | 0.2119 | 0.1890 |
| 1CNN vs PE G | 0.7049 | 0.6298 | 0.5794 | 0.5655 | 0.5544 | 0.5525 | 0.5334 |
| PE Y vs PE G | 0.9061 | 0.9511 | 0.9561 | 0.9352 | 0.8933 | 0.8817 | 0.9070 |
|  | 80% | 90% | 100% | 110% | 120% | 130% | 140% |
| 1KNN vs PE Y | 0.0297 | 0.0043 | -0.0021 | -0.0149 | -0.0214 | -0.0214 | -0.0214 |
| 1CNN vs PE Y | 0.4617 | 0.4503 | 0.4475 | 0.4418 | 0.4390 | 0.4390 | 0.4390 |
| 1KNN vs PE G | 0.1401 | 0.0933 | 0.0533 | 0.0360 | 0.0297 | 0.0297 | 0.0297 |
| 1CNN vs PE G | 0.5081 | 0.4829 | 0.4697 | 0.4645 | 0.4617 | 0.4617 | 0.4617 |
| PE Y vs PE G | 0.9275 | 0.9418 | 0.9668 | 0.9731 | 0.9730 | 0.9730 | 0.9730 |
|  | 150% | 160% | 170% | 180% | 190% | 200% |  |
| 1KNN vs PE Y | -0.0214 | -0.0214 | -0.0251 | -0.0251 | -0.0251 | -0.0251 |  |
| 1CNN vs PE Y | 0.4390 | 0.4390 | 0.4428 | 0.4428 | 0.4428 | 0.4428 |  |
| 1KNN vs PE G | 0.0107 | -0.0021 | -0.0021 | -0.0021 | -0.0021 | -0.0021 |  |
| 1CNN vs PE G | 0.4532 | 0.4475 | 0.4475 | 0.4475 | 0.4475 | 0.4475 |  |
| PE Y vs PE G | 0.9829 | 0.9897 | 0.9794 | 0.9794 | 0.9794 | 0.9794 |  |

[Table 3.4] Table of results using McNemar's test with '0' means the null hypothesis is accepted (two classifiers have the same error rate) and '1' means the alternative hypothesis is accepted (two classifiers have different error rates) for Transfusion database.

|  | 10% | 20% | 30% | 40% | 50% | 60% | 70% | 80% | 90% | 100% |
|---|---|---|---|---|---|---|---|---|---|---|
| 1KNN vs PEY | 1 | 1 | 1 | 1 | 1 | 1 | 1 | 1 | 1 | 1 |
| 1CNN vs PEY | 1 | 1 | 1 | 1 | 1 | 1 | 1 | 1 | 1 | 0 |
| 1KNN vs PEG | 1 | 1 | 1 | 1 | 1 | 1 | 1 | 1 | 1 | 1 |
| 1CNN vs PEG | 0 | 0 | 1 | 1 | 1 | 1 | 1 | 1 | 1 | 1 |
| PEY vs PEG | 1 | 1 | 0 | 0 | 0 | 0 | 0 | 0 | 0 | 0 |
|  | 110% | 120% | 130% | 140% | 150% | 160% | 170% | 180% | 190% | 200% |
| 1KNN vs PEY | 1 | 1 | 1 | 1 | 1 | 1 | 1 | 1 | 1 | 1 |
| 1CNN vs PEY | 0 | 0 | 0 | 0 | 0 | 0 | 0 | 0 | 0 | 0 |
| 1KNN vs PEG | 1 | 1 | 1 | 1 | 1 | 1 | 1 | 1 | 1 | 1 |
| 1CNN vs PEG | 1 | 1 | 1 | 1 | 1 | 0 | 0 | 0 | 0 | 0 |
| PEY vs PEG | 0 | 0 | 0 | 0 | 0 | 0 | 0 | 0 | 0 | 0 |